\documentclass{article}
\usepackage[T1]{fontenc}
\usepackage[utf8]{inputenc}
\usepackage{times}
\usepackage[margin=1in]{geometry}
\usepackage[round]{natbib}
\usepackage{amsmath,amssymb,booktabs,longtable,array,ragged2e,graphicx,tabularx}
\usepackage{hyperref}
\hypersetup{hidelinks}
\title{PINNMorph: Evolving Online Adaptation Policies for Physics-Informed Neural Networks}
\author{
Xu Yang$^{1}$,
Mingyang Yu$^{2}$,
Jun Zhang$^{3}$,
Keqian Li$^{1}$,
Jing Xu$^{2}$\\[0.5em]
$^{1}$East China Normal University\\
$^{2}$Nankai University\\
$^{3}$Hanyang University
}
\date{}
\begin{document}

\maketitle

\begin{abstract}
Physics-informed neural networks (PINNs) provide a learning-based framework for solving partial differential equations (PDEs), yet their training behavior can change substantially throughout optimization. Residual distributions, gradient interactions, regional learning difficulty, and model-capacity requirements may evolve over time, while the network architecture and major training mechanisms are typically determined before training. We propose PINNMorph, an online PINN adaptation framework based on large language model (LLM)-guided policy evolution. PINNMorph maintains a population of state-conditioned adaptation policies that map execution diagnostics to controlled interventions over topology modification, additive representation augmentation, objective balancing, gradient handling, adaptive sampling, and optimizer-phase control. At each intervention opportunity, candidate programs are instantiated from the current policy population, selected according to the observed training state, and applied directly to the PINN under training. The resulting model inherits its existing parameters and training state and continues optimization along the same trajectory. Execution outcomes are subsequently used to evaluate interventions and evolve the policy population. Unlike pre-training architecture search or fixed adaptation rules, PINNMorph jointly adapts the current PINN and the policies governing its interventions using feedback from actual training. Experiments on 13 PDE benchmarks show that PINNMorph achieves lower solution errors than SA-PINN, ConFIG, RoPINN, HARMONIC, and PINNsAgent across all evaluated problems. Ablation studies further examine the effects of online adaptation, state-conditioned intervention selection, and execution-feedback-driven policy evolution.
\end{abstract}

\section{Introduction}

Physics-informed neural networks (PINNs) solve partial differential equations (PDEs) by incorporating governing equations and boundary or initial conditions into neural-network training \citep{RAISSI2019686}. Despite their flexibility, PINNs remain highly sensitive to training dynamics. Imbalanced physical objectives, gradient conflicts, spectral bias, and non-uniform residual distributions can impair convergence and solution accuracy \citep{doi:10.1137/20M1318043,WANG2022110768,WU2023115671,krishnapriyan2021failure,rathore2024challenges}. Their effects also vary across PDEs and training regimes, as demonstrated by large-scale benchmarks such as PINNacle \citep{NEURIPS2024_8c63299f}.

A further difficulty is that the dominant training bottleneck can change \emph{within the same optimization trajectory}. Early training may be limited by imbalance among physical constraints, while later stages can suffer from gradient conflicts, localized residual concentration, insufficient sampling, or limited representational capacity \citep{pmlr-v202-daw23a,WANG2024116813}. Hence, the question is not only how a PINN should be adapted, but also what should be adapted as training evolves.

Existing adaptive PINN methods address specific aspects of this problem, including loss weighting, gradient coordination, adaptive sampling, and regional optimization \citep{MCCLENNY2023111722,ICLR2025_94e85561,WU2023115671,NEURIPS2024_c745bfa5}. However, the adaptation family is typically fixed in advance. Automated PINN design instead searches for suitable architectures or training configurations before training or across independent runs \citep{wang2023autopinn,pmlr-v267-wuwu25a}. Neither paradigm directly addresses state-dependent selection among multiple intervention families within the training trajectory of the same PINN.

We therefore formulate online PINN adaptation as a sequential decision problem. At each diagnosis point, the current physical error, optimization behavior, model state, and remaining budget determine whether intervention is needed and which type of change is appropriate. Unlike fixed adaptation rules or pre-training configuration selection, the intervention space spans heterogeneous and structured changes to model topology, representation, objectives, gradients, sampling, and optimization. This motivates using LLMs to instantiate state-conditioned interventions and evolve the policies that generate them.

We propose PINNMorph, an online PINN adaptation framework based on LLM-guided policy evolution. PINNMorph maintains a population of state-conditioned policies that map training diagnostics to interventions over topology modification, additive representation augmentation, objective balancing, gradient handling, adaptive sampling, and optimizer-phase control. At each diagnosis point, candidate actions are generated from the current state, and the selected intervention is applied directly to the PINN under optimization. Learned parameters and transferable training states are preserved whenever possible, allowing the model to adapt without restarting optimization. PINN design thus becomes a process coupled with training rather than a one-time choice made beforehand.

Evaluating an intervention introduces a second challenge. Continued training  may improve the PINN even without intervention, so post-intervention improvement alone does not reveal the effect of the action. PINNMorph therefore compares the realized trajectory with an estimated no-intervention continuation and uses the resulting execution feedback to evolve the policy population. Later decisions can thereby benefit from the observed consequences of earlier interventions along the same training trajectory.

We evaluate PINNMorph on 13 PDE benchmarks spanning nonlinear dynamics, multiscale systems, complex geometries, high-dimensional equations, and long-time problems. The comparison includes SA-PINN, ConFIG, RoPINN, HARMONIC, and PINNsAgent. PINNMorph achieves the lowest mean MSE on all 13 benchmarks, with a Friedman average rank of 1.00. Component ablations further examine the effects of online adaptation, state-conditioned action selection, and execution-feedback-driven policy evolution. Our main contributions are:

\begin{itemize}
    \item \textbf{Cross-family online adaptation.}
    We formulate PINN adaptation as a within-trajectory sequential decision problem in which the appropriate intervention family can change as training evolves.

    \item \textbf{State-conditioned policy evolution.}
    We introduce an LLM-guided population of policies that turns multi-view training diagnostics into structured interventions across model, representation, objective, gradient, sampling, and optimizer dimensions.

    \item \textbf{Execution-grounded feedback.}
    To distinguish intervention effects from improvements due to continued training, we evaluate each intervention relative to an estimated no-intervention continuation and feed the resulting state–action–outcome evidence back into policy evolution, enabling later decisions to learn from earlier interventions.
\end{itemize}

\section{Related Work}

\subsection{Physics-Informed Neural Networks and Adaptive Training}

PINNs formulate PDE solving as neural function approximation constrained by governing equations and boundary or initial conditions \citep{RAISSI2019686}. Subsequent work has investigated architectures and representations that improve trainability and expressiveness, including Fourier features \citep{WANG2021113938}, Transformer-based PINNs \citep{ICLR2024_a6f27630}, adaptive activation functions \citep{jagtap2020adaptive,zhangding2025activation}, causal training \citep{WANG2024116813}, residual-adaptive networks \citep{JMLR:v25:24-0313}, and domain-decomposed models \citep{JAGTAP2020113028,djagtap2020,KHARAZMI2021113547}. Geometry-aware trial functions can enforce boundary conditions exactly \citep{sukumar2022exact}, while gradient-enhanced residual objectives add derivative information to the PINN loss \citep{yu2022gradient}. PINNacle provides a broad benchmark showing that the relative effectiveness of such techniques varies substantially across PDE classes \citep{NEURIPS2024_8c63299f}.

A complementary line of research adapts the training process itself. SA-PINN learns point-wise loss weights \citep{MCCLENNY2023111722}; other methods dynamically balance physical objectives or transform their residuals \citep{cao2025wbpinn,gao2025adaptive,zhangdeng2025loss}. Residual-based and failure-informed methods adapt collocation distributions toward difficult regions \citep{WU2023115671,pmlr-v202-daw23a,doi:10.1137/22M1527763}; imbalanced-learning, random-walk, and causality-guided schemes offer further sampling strategies \citep{luo2025imbalanced,wanghu2025randomwalk,lin2025causality}. RoPINN extends optimization from individual points to adaptive local regions \citep{NEURIPS2024_c745bfa5}. Gradient-oriented approaches such as ConFIG and HARMONIC instead modify update directions to reduce conflicts among physical objectives \citep{ICLR2025_94e85561,ICLR2026_b5ffdfe4}. These approaches demonstrate the value of training-time adaptation, but each primarily operates within a predefined adaptation family. PINNMorph instead selects among multiple intervention families according to the current training state.

\subsection{Automated PINN Design}

Automated PINN design aims to reduce manual configuration by searching over architectures and training choices. Auto-PINN and NAS-PINN explore PDE-specific neural architectures through automated search \citep{wang2023autopinn,WANG2024112603}. Evolutionary approaches search PINN architectures and activation functions \citep{kaplarevic2023optimal,zhangyang2024evo}, while knowledge distillation has been used to discover network structures from trained models \citep{liu2025automatic}. More recently, PINNsAgent uses LLMs together with PDE information and experimental feedback to iteratively generate improved PINN configurations \citep{pmlr-v267-wuwu25a}, while Lang-PINN studies multi-agent language-model workflows for constructing physics-informed models from natural-language problem descriptions \citep{you2026langpinn}.

These methods generate candidate configurations and evaluate them as separate models or runs. PINNMorph instead operates \emph{inside} one ongoing training process: policies observe the evolving state of the current PINN and directly modify that model without restarting optimization.

\subsection{LLM-Driven Algorithm Evolution}

LLMs have also been used as generators within iterative algorithm-design loops, as surveyed by \citet{wu2025survey,liu2026survey}. OPRO uses previously evaluated solutions to guide later language-model proposals \citep{ICLR2024_3339f19c}, while FunSearch combines program generation with external execution-based evaluation \citep{RomeraParedes2024}. EoH evolves heuristic ideas and executable implementations through LLM generation and execution feedback \citep{pmlr-v235-liu24bs}; ReEvo incorporates reflective feedback during evolution \citep{NEURIPS2024_4ced59d4}; and LLaMEA iteratively improves optimization algorithms using runtime performance \citep{van2025llamea}. Subsequent work studies in-loop hyperparameter optimization, controlled mutations, and multiple design objectives within LLM-based heuristic evolution \citep{vanstein2025intheloop,yin2025mutation,yao2025moeoh}.

PINNMorph adopts the general principle of execution-feedback-driven evolution but changes the evolutionary object. Rather than evolving a standalone algorithm or independently evaluated model, PINNMorph evolves a state-conditioned policy \(\pi: s_t \mapsto a_t\), where $s_t$ is the current PINN training state and $a_t$ is an online adaptation action. Policy evolution and PINN optimization therefore interact within the same continuous training trajectory.

\section{Method}

\subsection{Problem Formulation}

Consider a general PDE defined on a computational domain $\Omega$, subject to boundary conditions and, for time-dependent problems, initial conditions:
\begin{equation}
\mathcal{F}[u](\mathbf{x}) = 0,\quad \mathbf{x}\in\Omega,
\qquad
\mathcal{B}[u](\mathbf{x}) = 0,\quad \mathbf{x}\in\partial\Omega,
\qquad
\mathcal{I}[u](\mathbf{x}) = 0,\quad \mathbf{x}\in\Omega_0.
\label{eq:pde_system}
\end{equation}

Here, $\mathbf{x}$ denotes the spatial or spatiotemporal coordinate, $u$ is the unknown physical field, $\Omega$ is the computational domain, $\partial\Omega$ denotes its boundary, and $\Omega_0$ denotes the initial-time domain. The operators $\mathcal{F}$, $\mathcal{B}$, and $\mathcal{I}$ represent the governing equation, boundary condition, and initial condition, respectively. The initial-condition term is omitted when it is not required.

A PINN approximates the solution using a neural network $u_\theta(\mathbf{x})$ with parameters $\theta$ and minimizes
\begin{equation}
\mathcal{L}(\theta)
=
\lambda_{\mathrm{PDE}}\mathcal{L}_{\mathrm{PDE}}
+
\lambda_{\mathrm{BC}}\mathcal{L}_{\mathrm{BC}}
+
\lambda_{\mathrm{IC}}\mathcal{L}_{\mathrm{IC}},
\label{eq:pinn_loss}
\end{equation}
where $\mathcal{L}_{\mathrm{PDE}}$, $\mathcal{L}_{\mathrm{BC}}$, and $\mathcal{L}_{\mathrm{IC}}$ denote the governing-equation, boundary-condition, and initial-condition losses, respectively, and $\lambda_{\mathrm{PDE}}$, $\lambda_{\mathrm{BC}}$, and $\lambda_{\mathrm{IC}}$ are their corresponding weights.

\begin{figure}[t]
\centering
\includegraphics[width=\textwidth]{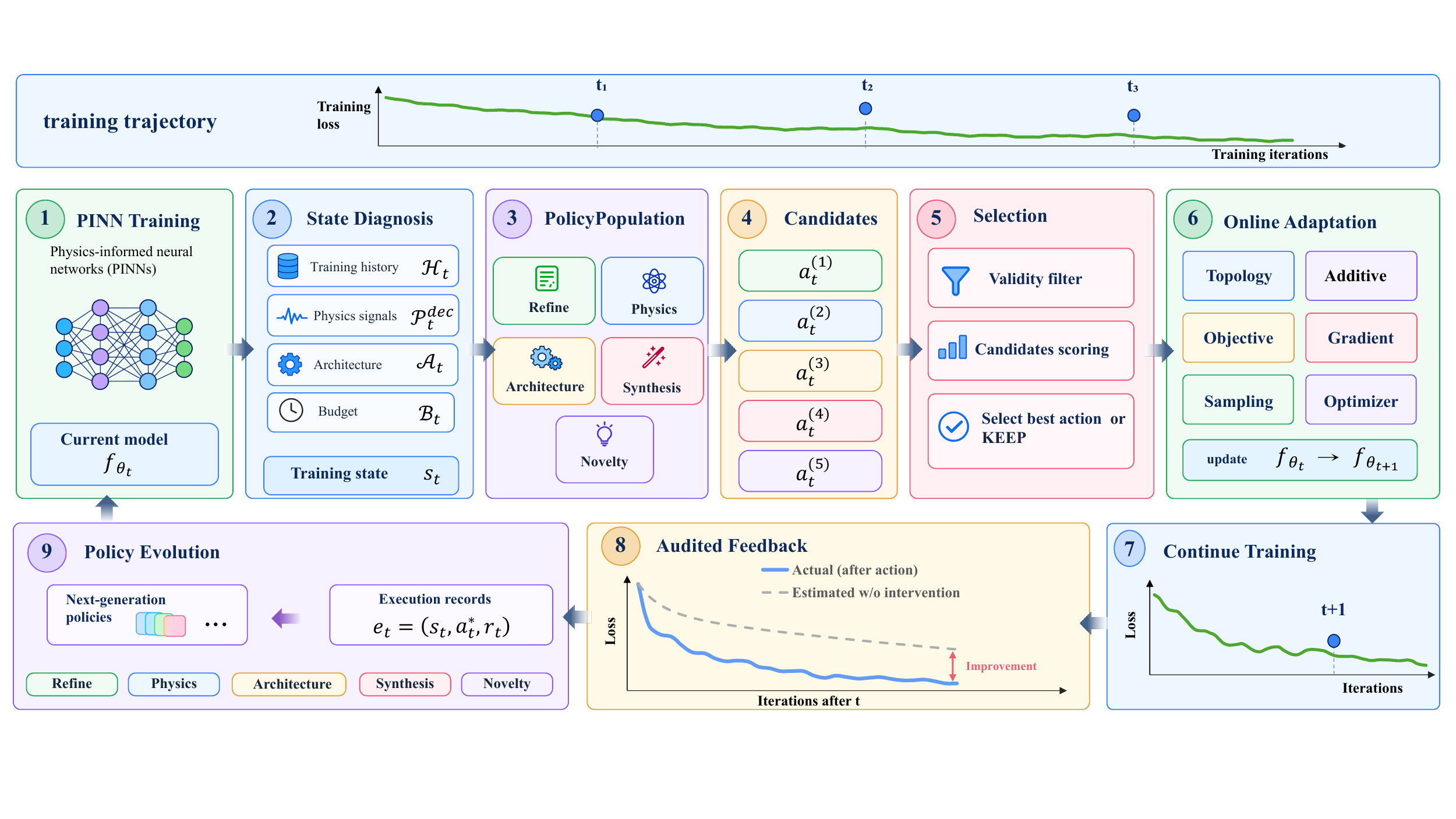}
\caption{
Overview of PINNMorph. At each adaptation point, PINNMorph diagnoses the current training state, uses a policy population to propose candidate actions, selects an executable action based on validity and state-conditioned scoring, and applies it online to the PINN under training. The resulting execution feedback is audited against the estimated no-intervention trajectory and used to evolve the policy population for subsequent adaptations.
}
\label{fig}
\end{figure}
\subsection{Method Overview}

PINNMorph maintains a PINN under optimization throughout training. Let $f_{\theta_t}^{A_t}$ denote the PINN at training step $t$, with parameters $\theta_t$ and architecture $A_t$. At each diagnosis point, PINNMorph constructs a training state
\begin{equation}
s_t=\mathcal{D}(\mathcal{H}_t,\mathcal{P}_t,A_t,B_t),
\label{eq:training_state}
\end{equation}
where $\mathcal{H}_t$ summarizes recent optimization and adaptation history, $\mathcal{P}_t$ contains physics diagnostics, and $B_t$ represents the remaining training budget and execution context.

PINNMorph maintains a population of \(K=5\) state-conditioned adaptation policies,
\begin{equation}
\Pi_g=\{\pi_g^{(1)},\ldots,\pi_g^{(K)}\},
\label{eq:policy_population}
\end{equation}
which generate candidate interventions from the current training state. The selected intervention is executed on the current PINN, and its subsequent execution feedback is used to update the policy population.

\subsection{Training-State Diagnosis}

PINNMorph uses two fixed held-out collocation sets: a decision set $\mathcal{B}_{\mathrm{dec}}$ and an audit set $\mathcal{B}_{\mathrm{aud}}$. Neither participates in gradient updates.

At diagnosis point $t$, $\mathcal{B}_{\mathrm{dec}}$ is used to estimate physics diagnostics, including PDE, boundary, and initial-condition residuals, residual concentration, dominant constraint violations, gradient statistics, and optimization stagnation. These quantities, together with architecture and execution history, form the state $s_t$.

The audit set $\mathcal{B}_{\mathrm{aud}}$ is reserved for evaluating the physical effect of an executed intervention. Separating the decision and audit sets reduces direct coupling between action selection and feedback estimation.

\subsection{Adaptation Policies}

Each policy maps the current training state to an adaptation action, \(a_t^{(k)}=\pi_g^{(k)}(s_t)\). The initial policy population is generated from the PDE description, current state, architecture, and controlled action space. Subsequent generations use five complementary modes: Refinement, Physics-Guided, Architecture-Guided, Synthesis, and Novelty.

These modes respectively improve effective policies, respond to dominant physical errors, adjust representational capacity, combine useful mechanisms, and explore new adaptation directions.

\subsection{Online Adaptation}

The intervention space comprises six action families: topology modification, additive representation, objective balancing, gradient coordination, adaptive sampling, and optimizer-phase control. At each diagnosis point, the policy population generates multiple candidate actions. PINNMorph scores these candidates based on the current training bottleneck and historical execution outcomes, and selects the most appropriate action:

\begin{equation}
a_t^\ast
=
\arg\max_{a\in\mathcal{C}_t}
\mathcal{S}_\phi
\left(
s_t,
a,
\mathcal{H}_t^{\mathrm{exec}}
\right).
\label{eq:action_selection}
\end{equation}

The scoring function favors actions aligned with the current training bottleneck while incorporating evidence from previous interventions. If no action is sufficiently appropriate, training continues without intervention.

The selected action is applied directly to the current PINN. Structural interventions preserve learned parameters and transferable optimizer states whenever possible, while non-structural actions modify the corresponding training mechanism. At most one action is executed at each diagnosis point. Detailed action mechanisms and execution rules are provided in Appendix C.

\subsection{Policy Evolution}

After executing $a_t^\ast$, PINNMorph evaluates its effect from the subsequent training trajectory. Because ordinary optimization would also change the residual without intervention, the observed trajectory is compared with an estimated continuation trend:

\begin{equation}
r_t
=
\mathcal{U}
\left(
\mathcal{P}_{>t},
\widehat{\mathcal{P}}_{>t}^{\,0}
\right).
\label{eq:execution_feedback}
\end{equation}

Here, $\mathcal{P}_{>t}$ is the observed post-intervention trajectory, $\widehat{\mathcal{P}}_{>t}^{\,0}$ the estimated no-intervention continuation, and $\mathcal{U}$ the utility function comparing the two. The resulting feedback \(r_t\) estimates the incremental effect of the intervention. All intervention outcomes are recorded in the execution history. Numerical policy credit is assigned only when the decision and audit sets meet the forecast-eligibility and cross-set agreement criteria; consistent but low-confidence outcomes may provide bounded qualitative evidence for subsequent policy evolution.

The next policy population is generated as
\begin{equation}
\Pi_{g+1}
=
\mathcal{E}_\psi
\left(
\Pi_g,
\mathcal{H}_g^{\mathrm{exec}},
s_t
\right).
\label{eq:policy_evolution}
\end{equation}
Policies associated with beneficial interventions provide positive experience for subsequent generations, whereas ineffective or narrowly applicable actions constrain future updates. In this way, training-state diagnosis, online adaptation, and feedback-driven policy evolution form a closed loop. Detailed continuation estimation and execution-feedback rules are provided in Appendix D.
\section{Experiments}

We evaluate PINNMorph on 13 PDEs with diverse training characteristics. The experiments focus on overall solution accuracy, the contribution of online adaptation, the role of state-conditioned action selection, and the effect of execution-feedback-driven policy evolution.

\subsection{Experimental Setup}

\textbf{Benchmarks.} We evaluate PINNMorph on $13$ PDEs covering nonlinear transport, multiscale dynamics, complex geometries, high-dimensional equations, and long-time dynamics. For PINNacle tasks, we follow the public problem definitions and reference-solution settings \citep{NEURIPS2024_8c63299f}. Full PDE specifications are provided in Appendix A.

\textbf{Baselines.} We compare with SA-PINN \citep{MCCLENNY2023111722}, ConFIG \citep{ICLR2025_94e85561}, RoPINN \citep{NEURIPS2024_c745bfa5}, HARMONIC \citep{ICLR2026_b5ffdfe4}, and PINNsAgent \citep{pmlr-v267-wuwu25a}, covering adaptive weighting, gradient coordination, regional optimization, and LLM-based PINN design. 

\textbf{Metrics.} We use mean squared error (MSE) as the primary metric:
\begin{equation}
\mathrm{MSE}
=
\frac{1}{Nc}
\sum_{i=1}^{N}
\left\|
\mathbf{u}_{\theta}(\mathbf{x}_i)
-
\mathbf{u}^{*}(\mathbf{x}_i)
\right\|_2^2.
\label{eq:mse}
\end{equation}
Here, $\mathbf{u}_{\theta}(\mathbf{x}_i)$ is the solution obtained
from the trained PINN, and $\mathbf{u}^{*}(\mathbf{x}_i)$ is the
reference solution at evaluation point $\mathbf{x}_i$. The factor
$Nc$ is the total number of scalar solution values in the evaluation
set. All methods are evaluated over five random seeds, and we report
mean $\pm$ standard deviation.

\textbf{Implementation Details.} All methods use a fixed training budget of $20{,}000$ optimizer steps. DeepSeek-V4-Flash initializes and evolves the policy population, while Qwen3.8-27B evaluates candidate actions. PINNsAgent also uses DeepSeek-V4-Flash. The same LLM configuration is used across all PDEs. Additional implementation details are provided in Appendix B.

\subsection{Main Results}

Table~\ref{tab:main-results} compares the final MSE of PINNMorph and the baselines on all $13$ PDEs. Each method is independently evaluated with five random seeds, and results are reported as mean $\pm$ standard deviation.
\begin{table}[t]
\centering
\caption{Solution MSE on $13$ PDE benchmarks. Results are reported as mean $\pm$ standard deviation over five independent runs. Lower is better.}
\label{tab:main-results}

\resizebox{\linewidth}{!}{%
\begin{tabular}{ccccccc}
\toprule
\textbf{PDE}
& \textbf{SA-PINN}
& \textbf{ConFIG}
& \textbf{RoPINN}
& \textbf{HARMONIC}
& \textbf{PINNsAgent}
& \textbf{PINNMorph} \\
\midrule

KS 1D
& 6.40E+0 $\pm$ 4.63E-2
& 1.47E+1 $\pm$ 4.60E-1
& 6.37E+0 $\pm$ 5.01E-1
& 5.55E+1 $\pm$ 5.88E+0
& 1.12E+0 $\pm$ 3.27E-2
& \textbf{1.05E+0 $\pm$ 3.44E-1} \\

Burgers 1D
& 4.77E-4 $\pm$ 3.08E-5
& 2.47E-4 $\pm$ 1.27E-4
& 6.31E-4 $\pm$ 6.35E-5
& 1.32E-4 $\pm$ 2.13E-5
& 7.24E-5 $\pm$ 1.49E-5
& \textbf{6.48E-5 $\pm$ 3.29E-5} \\

NS 2D CG
& 7.10E-3 $\pm$ 3.11E-4
& 2.46E-2 $\pm$ 7.81E-3
& 9.91E-3 $\pm$ 5.27E-3
& 5.34E-3 $\pm$ 4.84E-4
& 4.44E-3 $\pm$ 1.30E-3
& \textbf{1.47E-3 $\pm$ 4.13E-4} \\

Heat 2D CG
& 3.41E-3 $\pm$ 6.07E-5
& 1.60E-2 $\pm$ 3.96E-3
& 4.92E-3 $\pm$ 8.93E-4
& 2.87E-2 $\pm$ 4.96E-4
& 3.61E-3 $\pm$ 8.71E-4
& \textbf{1.81E-3 $\pm$ 4.48E-4} \\

Heat 2D MS
& 3.03E-4 $\pm$ 2.25E-5
& 3.18E-4 $\pm$ 2.99E-5
& 3.95E-4 $\pm$ 8.06E-6
& 4.20E-4 $\pm$ 6.47E-5
& 3.13E-4 $\pm$ 2.59E-5
& \textbf{2.34E-4 $\pm$ 2.21E-5} \\

Heat 2D VC
& 1.45E-2 $\pm$ 5.17E-3
& 3.52E-2 $\pm$ 4.91E-3
& 4.70E-1 $\pm$ 4.86E-3
& 9.95E-2 $\pm$ 3.86E-2
& 5.62E-3 $\pm$ 3.33E-3
& \textbf{1.82E-3 $\pm$ 2.54E-4} \\

Burgers 2D
& 5.13E-1 $\pm$ 5.07E-2
& 4.44E+0 $\pm$ 1.05E+0
& 3.43E-1 $\pm$ 3.16E-2
& 4.48E+0 $\pm$ 6.05E-1
& 2.13E-1 $\pm$ 1.46E-2
& \textbf{1.75E-1 $\pm$ 4.14E-2} \\

Poisson 2D C
& 4.40E-1 $\pm$ 5.45E-2
& 1.17E+0 $\pm$ 4.11E-2
& 1.46E-1 $\pm$ 6.66E-2
& 3.50E+0 $\pm$ 5.72E-1
& 5.43E-1 $\pm$ 1.22E-1
& \textbf{1.01E-1 $\pm$ 3.75E-3} \\

Poisson 2D CG
& 1.72E-1 $\pm$ 4.53E-2
& 4.69E-1 $\pm$ 1.95E-1
& 1.65E+0 $\pm$ 7.80E-1
& 1.24E-1 $\pm$ 3.99E-2
& 7.68E-2 $\pm$ 8.19E-3
& \textbf{2.62E-2 $\pm$ 1.09E-2} \\

Poisson 2D MA
& 6.81E+0 $\pm$ 5.82E+0
& 3.31E+1 $\pm$ 9.97E+0
& 3.18E+0 $\pm$ 3.86E-1
& 5.64E+1 $\pm$ 4.99E+1
& 4.16E+0 $\pm$ 8.98E-1
& \textbf{2.65E+0 $\pm$ 7.99E-1} \\

Gray--Scott 2D
& 1.81E-2 $\pm$ 2.52E-3
& 1.15E-1 $\pm$ 8.65E-3
& 9.96E-3 $\pm$ 4.70E-4
& 7.43E-2 $\pm$ 1.06E-2
& 9.02E-3 $\pm$ 2.84E-4
& \textbf{3.34E-3 $\pm$ 4.74E-4} \\

Poisson 3D CG
& 3.05E-2 $\pm$ 6.19E-4
& 1.14E-1 $\pm$ 7.10E-2
& 2.07E-1 $\pm$ 1.46E-2
& 7.54E-2 $\pm$ 6.16E-3
& 1.74E-2 $\pm$ 1.17E-2
& \textbf{6.79E-3 $\pm$ 1.23E-3} \\

Poisson 5D
& 5.84E-6 $\pm$ 3.62E-7
& 2.13E-6 $\pm$ 1.42E-7
& 7.41E-7 $\pm$ 8.02E-8
& 3.94E-6 $\pm$ 2.77E-6
& 2.32E-6 $\pm$ 1.26E-6
& \textbf{6.26E-7 $\pm$ 8.32E-8} \\

\midrule
\textbf{Friedman Avg. Rank}
& \multicolumn{1}{c}{3.69}
& \multicolumn{1}{c}{4.77}
& \multicolumn{1}{c}{4.08}
& \multicolumn{1}{c}{4.92}
& \multicolumn{1}{c}{2.54}
& \multicolumn{1}{c}{\textbf{1.00}} \\

\bottomrule
\end{tabular}%
}
\end{table}

PINNMorph achieves the lowest mean MSE on all $13$ PDE benchmarks, with a Friedman average rank of $1.00$, indicating consistently strong performance across the benchmark suite.

The improvements are substantial on several representative problems. Compared with the best-performing baseline on each PDE, PINNMorph reduces MSE by $67.6\%$ on Heat 2D VC, $66.9\%$ on NS 2D CG, $65.9\%$ on Poisson 2D CG, $63.0\%$ on Gray--Scott 2D, and $61.0\%$ on Poisson 3D CG. These gains span PDEs with different dimensionalities, geometries, coefficients, and nonlinear dynamics.

The advantage also persists on tasks where the performance margins are smaller. PINNMorph achieves the best mean MSE on Burgers 1D, Heat 2D MS, and Poisson 5D, where the strongest baselines already attain relatively low errors. Taken together, the results show that the improvement is consistent across diverse PDE settings rather than driven by a small subset of benchmarks.

The contributions of online adaptation, state-conditioned action selection, and policy evolution are further examined through ablation studies.

\subsection{Component Ablation}
To analyze the three core mechanisms of PINNMorph, we construct three ablated variants: No Morph, Random Selection, and Random Evolution. All variants use the same base PINN, random seeds, training data, and 20,000-step optimization budget as the complete method. Each variant is independently run five times, and MSE is reported as mean ± standard deviation. No Morph: Online adaptation is removed. The base PINN retains a fixed model structure and fixed training mechanisms throughout optimization. Random Selection: The dependence of final action selection on the training state $s_t$ is removed. At each adaptation point, the final action is sampled randomly from the executable candidate set. Random Evolution: The generation of the next population does not use execution feedback from previous actions; instead, the evolution direction is randomized. Table~\ref{tab:ablation-config} summarizes the component configurations. Table~\ref{tab:ablation-results} reports the final MSE of the three ablated variants and the complete PINNMorph on the 13 PDEs.

\begin{table}[t]
\caption{Component configurations used in the ablation study.}
\label{tab:ablation-config}
\centering
\resizebox{\textwidth}{!}{%
\begin{tabular}{ccccc}
\toprule
\textbf{Variant} & \textbf{Online Adaptation} & \textbf{State-Conditioned Selection} & \textbf{Policy Evolution} & \textbf{Feedback Evolution} \\
\midrule
No Morph & $\times$ & $\times$ & $\times$ & $\times$ \\
Random Selection & \checkmark & $\times$ & \checkmark & \checkmark \\
Random Evolution & \checkmark & \checkmark & \checkmark & $\times$ \\
\textbf{PINNMorph} & \checkmark & \checkmark & \checkmark & \checkmark \\
\bottomrule
\end{tabular}
}
\end{table}

\begin{table}[!t]
\caption{Component ablation results. Results are reported as MSE (mean \(\pm\) standard deviation) over five independent runs. Lower is better.}
\label{tab:ablation-results}
\centering
\resizebox{\textwidth}{!}{%
\begin{tabular}{ccccc}
\toprule
\textbf{PDE} & \textbf{No Morph} & \textbf{Random Selection} & \textbf{Random Evolution} & \textbf{\textbf{PINNMorph}} \\
\midrule
KS 1D
& 5.21E+0 $\pm$ 1.76E+0
& 6.24E+0 $\pm$ 1.87E+0
& 5.03E+0 $\pm$ 5.74E-1
& \textbf{1.05E+0 $\pm$ 3.44E-1} \\

Burgers 1D
& 6.82E-4 $\pm$ 5.05E-4
& 3.62E-4 $\pm$ 1.35E-4
& 7.92E-4 $\pm$ 4.96E-4
& \textbf{6.48E-5 $\pm$ 3.29E-5} \\

NS 2D CG
& 8.72E-3 $\pm$ 3.38E-3
& 9.46E-2 $\pm$ 1.10E-2
& 4.39E-3 $\pm$ 1.86E-3
& \textbf{1.47E-3 $\pm$ 4.13E-4} \\

Heat 2D CG
& 5.17E-3 $\pm$ 6.43E-4
& 1.95E-2 $\pm$ 4.83E-3
& 4.01E-2 $\pm$ 5.46E-3
& \textbf{1.81E-3 $\pm$ 4.48E-4} \\

Heat 2D MS
& 4.26E-3 $\pm$ 3.91E-4
& 3.89E-3 $\pm$ 2.72E-4
& 4.43E-3 $\pm$ 8.00E-5
& \textbf{2.34E-4 $\pm$ 2.21E-5} \\

Heat 2D VC
& 3.43E-2 $\pm$ 1.67E-2
& 4.20E-2 $\pm$ 1.46E-2
& 4.94E-3 $\pm$ 2.07E-3
& \textbf{1.82E-3 $\pm$ 2.54E-4} \\

Burgers 2D
& 4.26E+0 $\pm$ 5.20E-1
& 5.51E+1 $\pm$ 1.54E+1
& 2.76E+0 $\pm$ 1.62E-1
& \textbf{1.75E-1 $\pm$ 4.14E-2} \\

Poisson 2D C
& 6.26E+0 $\pm$ 8.75E-1
& 7.43E+0 $\pm$ 4.02E-1
& 1.04E+0 $\pm$ 3.74E-1
& \textbf{1.01E-1 $\pm$ 3.75E-3} \\

Poisson 2D CG
& 3.77E-1 $\pm$ 1.67E-1
& 5.08E-1 $\pm$ 2.53E-1
& 7.50E-1 $\pm$ 1.02E-1
& \textbf{2.62E-2 $\pm$ 1.09E-2} \\

Poisson 2D MA
& 1.30E+1 $\pm$ 6.09E+0
& 1.70E+1 $\pm$ 6.12E+0
& 1.04E+1 $\pm$ 2.35E+0
& \textbf{2.65E+0 $\pm$ 7.99E-1} \\

Gray--Scott 2D
& 3.31E-2 $\pm$ 1.21E-2
& 2.51E-2 $\pm$ 1.14E-2
& 7.35E-2 $\pm$ 3.00E-3
& \textbf{3.34E-3 $\pm$ 4.74E-4} \\

Poisson 3D CG
& 8.59E-2 $\pm$ 3.32E-3
& 6.69E-2 $\pm$ 1.05E-2
& 1.56E-2 $\pm$ 6.10E-3
& \textbf{6.79E-3 $\pm$ 1.23E-3} \\

Poisson 5D
& 9.00E-6 $\pm$ 3.72E-6
& 2.95E-6 $\pm$ 8.96E-7
& 2.80E-6 $\pm$ 5.12E-7
& \textbf{6.26E-7 $\pm$ 8.32E-8} \\

\bottomrule
\end{tabular}
}
\end{table}

Because the MSE values span several orders of magnitude across PDEs, we compute the error ratio of each ablated variant relative to the complete PINNMorph:

\begin{equation}
r_p = \frac{\mathrm{MSE}_{\mathrm{ablation},p}}{\mathrm{MSE}_{\mathrm{PINNMorph},p}}.
\label{eq:error_ratio}
\end{equation}

We then summarize performance across PDEs using \(r_{\mathrm{median}} = \operatorname{median}_{p}\left(r_p\right)\). A value of \(r_{\mathrm{median}}>1\) indicates that the ablated variant has a higher overall solution error than PINNMorph.

The full PINNMorph achieves the lowest mean MSE across all $13$ PDEs. As shown in Table~\ref{tab:cost_and_ratio}, Removing online adaptation increases the median error ratio to $12.65\times$, demonstrating the importance of adapting the PINN during optimization. Random Selection yields a median error ratio of $10.77\times$ and is inferior to No Morph on $8$ of $13$ tasks. Although random interventions can occasionally improve performance substantially on individual PDEs, their degradation on the majority of tasks indicates that online adaptation alone does not guarantee consistent improvements when intervention selection is decoupled from the current training state. Random Evolution achieves the lowest median error ratio among the three ablations at $10.30\times$ and is the strongest ablation on $8$ of $13$ tasks, indicating that state diagnosis and state-conditioned action selection already provide substantial benefits even without execution-feedback-driven evolution. Nevertheless, the complete PINNMorph further improves performance on all $13$ PDEs, reducing the median error by approximately one order of magnitude relative to each ablated variant. These results demonstrate the complementary contributions of online adaptation, state-conditioned selection, and execution-feedback-driven policy evolution.

\subsection{Analysis of the Online Adaptation Process}

\begin{figure}[t]
    \centering
    \includegraphics[width=\textwidth]{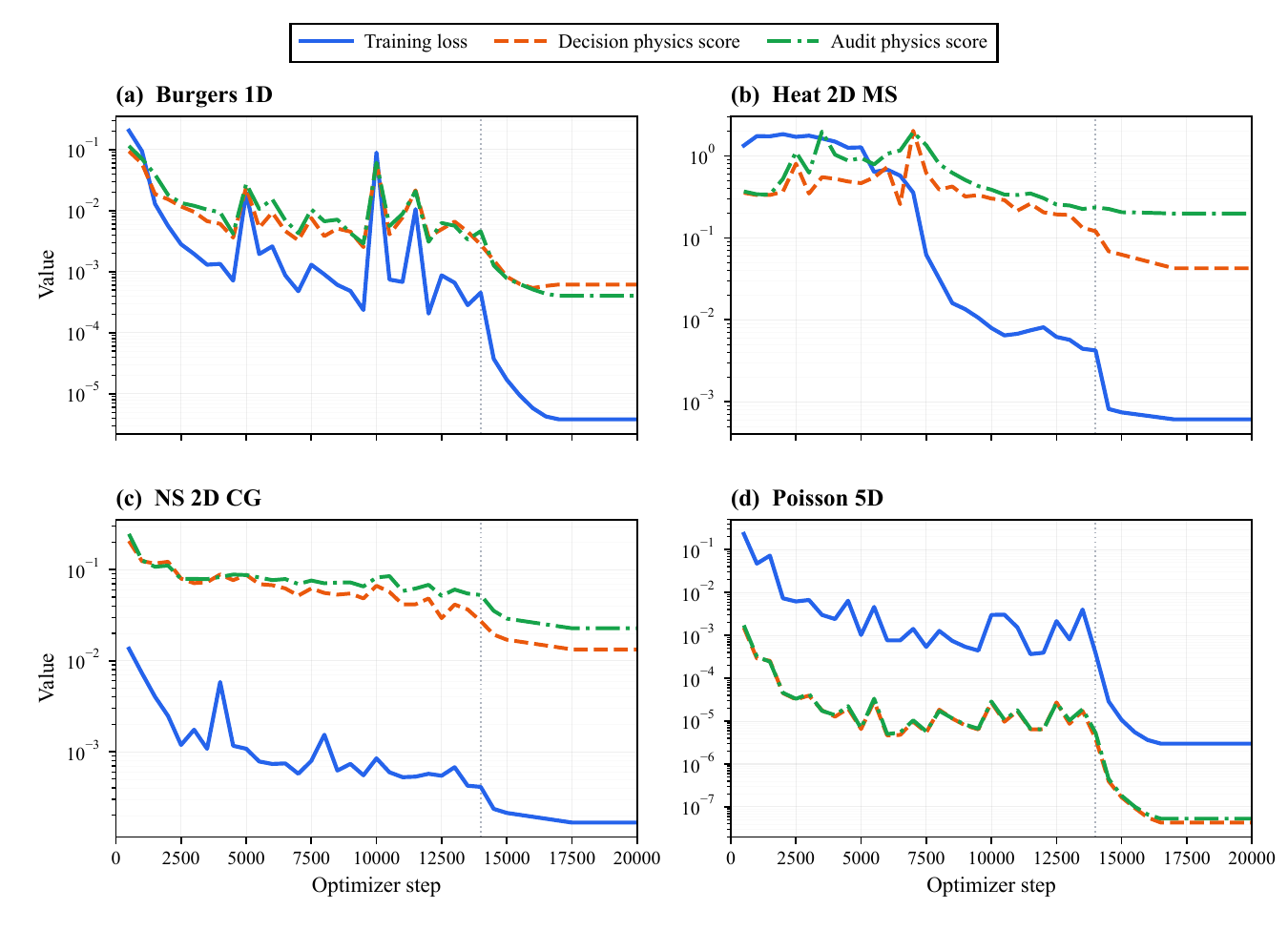}
\caption{
Representative training trajectories of PINNMorph. Training loss and the Decision and Audit physics scores are tracked throughout optimization, with the two physics scores computed on the fixed decision and audit sets, respectively.
}
    \label{fig:online_trajectories}
\end{figure}

To examine the online behavior of PINNMorph, we analyze the evolution of optimization and physics-related signals together with representative adaptations. Figure~\ref{fig:online_trajectories} shows the training loss, Decision physics score, and Audit physics score over four representative runs. The Decision physics score reflects the physical state available to online decision making, while the Audit physics score is computed on the fixed audit set $\mathcal{B}_{\mathrm{aud}}$. Both are residual-based quantities, with lower values indicating better satisfaction of the governing equations.

As shown in Figure~\ref{fig:online_trajectories}, these signals evolve non-uniformly and may improve at different rates, indicating that the dominant bottleneck can shift during optimization. Training loss alone therefore cannot distinguish whether poor progress arises from spatial error concentration, limited representation, or interactions among physical objectives. This motivates state-dependent diagnostics for online adaptation.

Table~\ref{tab:online_adaptation_cases} summarizes representative adaptations under different diagnosed training states. The value before intervention is measured at the intervention point, while the value after intervention is recorded 500 iterations later. Different bottlenecks favor different responses: representation-related difficulties can trigger localized specialization or global feature augmentation, spatial imbalance favors adaptive sampling, and conflicts among physical objectives can motivate structural specialization. Similar symptoms, such as slow progress or a plateau, may therefore require different interventions depending on the diagnosed state.

PINNMorph applies the selected action directly to the current PINN and evaluates its effect during continued optimization. The resulting execution feedback is then used for policy evolution, closing the loop between state diagnosis, intervention, execution, and subsequent adaptation.

\begin{table}[!t]
\centering
\caption{Representative online adaptations of PINNMorph and their effects under different training states.}
\label{tab:online_adaptation_cases}

\small
\setlength{\tabcolsep}{5pt}
\renewcommand{\arraystretch}{1.12}

\begin{tabular}{cccc}
\toprule
\textbf{Training state} &
\textbf{Diagnostic signal} &
\textbf{Action} &
\textbf{Before $\rightarrow$ After} \\
\midrule
Localized error &
Shock-localized residual &
Localized Expert &
$1.01\times10^{-2} \rightarrow 6.32\times10^{-4}$ \\

Representation limit &
Non-local residual &
Fourier Branch &
$3.71\times10^{-3} \rightarrow 2.59\times10^{-4}$ \\

Spatial imbalance &
Residual imbalance &
Adaptive Sampling &
$3.08\times10^{-2} \rightarrow 7.63\times10^{-3}$ \\

Gradient conflict &
Inter-equation conflict &
Topology Split &
$1.36\times10^{-3} \rightarrow 6.53\times10^{-4}$ \\

Spatial imbalance &
High-D residual imbalance &
Adaptive Sampling &
$2.26\times10^{-3} \rightarrow 8.08\times10^{-4}$ \\

Optimization plateau &
Stable gradients &
Correction Branch &
$4.86\times10^{-3} \rightarrow 1.46\times10^{-3}$ \\
\bottomrule
\end{tabular}
\end{table}

\subsection{Computational Cost}

Table~\ref{tab:cost_and_ratio} shows that PINNMorph introduces only moderate wall-clock overhead under the same optimization budget. This is because it adapts the current PINN in place rather than restarting training with a new model. The additional cost mainly comes from periodic diagnosis, LLM inference, intervention execution, and policy updates. Compared with No Morph, PINNMorph incurs approximately \(4.6\%\) additional wall-clock time while achieving substantially lower solution error.

\begin{table}[t]
\centering
\caption{Computational cost and median MSE ratios of PINNMorph and its ablated variants. All variants use the same $20{,}000$-step optimization budget. Median MSE ratios are computed relative to PINNMorph.}
\label{tab:cost_and_ratio}
\small
\begin{tabular}{lcccc}
\toprule
\textbf{Variant}
& \textbf{LLM Calls}
& \textbf{LLM Tokens}
& \textbf{Wall Time (s)}
& \textbf{Median MSE Ratio} \\
\midrule

No Morph
& 0
& 0
& 1250
& 12.65$\times$ \\

Random Selection
& 30
& $6.52\times10^{4}$
& 1273
& 10.77$\times$ \\

Random Evolution
& 33
& $1.06\times10^{5}$
& 1286
& 10.30$\times$ \\

PINNMorph
& 38
& $4.40\times10^{5}$
& 1308
& 1.00$\times$ \\

\bottomrule
\end{tabular}
\end{table}

\section{Conclusion}

We presented PINNMorph, an online adaptation framework for physics-informed neural networks that integrates training-state diagnosis, state-conditioned intervention, and execution-feedback-driven policy evolution within a single training trajectory. Rather than fixing an adaptation mechanism in advance, PINNMorph selects among heterogeneous intervention families according to the evolving training state, applies the selected action directly to the current model, and preserves learned parameters and transferable training state whenever possible. The resulting execution feedback is then used to improve subsequent adaptation policies. Across 13 PDE benchmarks, PINNMorph achieves the lowest mean MSE, with ablations confirming the value of online adaptation, state-conditioned selection, and feedback-driven policy evolution. With the same 20,000-step budget, it incurs only about 4.6\% wall-clock overhead over No Morph.

\appendix

\numberwithin{equation}{section}
\section{PDE Benchmark Specifications}
\label{app:pde_benchmarks}

\subsection{Burgers 1D}
\label{app:burgers1d}
For $(x,t)\in[-1,1]\times[0,1]$,
\begin{equation}
u_t+uu_x-\nu u_{xx}=0,\qquad \nu=\frac{0.01}{\pi},
\label{eq:appendix_burgers1d}
\end{equation}
with $u(x,0)=-\sin(\pi x)$ and $u(-1,t)=u(1,t)=0$.

\subsection{Burgers 2D}
\label{app:burgers2d}
For $\boldsymbol{u}=(u,v)$ on $[0,4]^2\times[0,1]$,
\begin{equation}
\partial_t\boldsymbol{u}
+(\boldsymbol{u}\cdot\nabla)\boldsymbol{u}
-10^{-3}\Delta\boldsymbol{u}=\boldsymbol{0}.
\label{eq:appendix_burgers2d}
\end{equation}
The initial field $\boldsymbol{u}(x,y,0)=\boldsymbol{u}_0(x,y)$ is prescribed.
Both components satisfy periodic value conditions on opposite spatial sides.

\subsection{Gray--Scott 2D}
\label{app:grayscott}
For $(x,y,t)\in[-1,1]^2\times[0,200]$,
\begin{equation}
\begin{aligned}
u_t&=\varepsilon_u\Delta u+b(1-u)-uv^2,\\
v_t&=\varepsilon_v\Delta v-dv+uv^2,
\end{aligned}
\qquad
(\varepsilon_u,\varepsilon_v)=(10^{-5},5\times10^{-6}),
\quad b=0.04,\quad d=0.1.
\label{eq:appendix_grayscott}
\end{equation}
The initial conditions are
\begin{align}
u(x,y,0)&=1-\exp\!\left[-80((x+0.05)^2+(y+0.02)^2)\right],\\
v(x,y,0)&=\exp\!\left[-80((x-0.05)^2+(y-0.02)^2)\right].
\end{align}

\subsection{Heat 2D CG}
\label{app:heatcomplex}
Let $\Omega=[-8,8]\times[-12,12]$ with disks removed. Eleven disks
have radius $1$ and centers
\begin{align*}
C_1=\{&(-4,-3),(4,-3),(-4,3),(4,3),(-4,-9),(4,-9),\\
&(-4,9),(4,9),(0,0),(0,6),(0,-6)\},
\end{align*}
and six disks have radius $0.4$ and centers

$$
C_{0.4}=\{(-3.2,-6),(-3.2,6),(3.2,-6),(3.2,6),(-3.2,0),(3.2,0)\}.
$$

For $t\in[0,3]$,
\begin{equation}
u_t-\Delta u=0,\qquad u(\boldsymbol{x},0)=0,
\qquad \partial_nu=c-u,
\label{eq:appendix_heatcomplex}
\end{equation}
where $c=5$ on radius-$1$ holes, $c=1$ on radius-$0.4$ holes,
and $c=0.1$ on the outer boundary.

\subsection{Heat 2D MS}
\label{app:heatmultiscale}
For $(x,y,t)\in[0,1]^2\times[0,5]$,
\begin{equation}
u_t-\kappa_xu_{xx}-\kappa_yu_{yy}=0,\qquad
\kappa_x=\frac{1}{(500\pi)^2},\quad \kappa_y=\frac{1}{\pi^2},
\label{eq:appendix_heatmultiscale}
\end{equation}
with $u(x,y,0)=\sin(20\pi x)\sin(\pi y)$ and
$u|_{\partial[0,1]^2}=0$. The analytic solution is
\begin{equation}
u^*(x,y,t)=\sin(20\pi x)\sin(\pi y)
\exp\!\left[-(\kappa_x(20\pi)^2+\kappa_y\pi^2)t\right].
\label{eq:appendix_heatmultiscale_exact}
\end{equation}

\subsection{Heat 2D VC}
\label{app:heatvarying}
For $(x,y,t)\in[0,1]^2\times[0,5]$,
\begin{equation}
u_t-a(x,y)\Delta u
-200\sin(\pi x)\sin(5\pi y)\sin(\pi t)=0,
\label{eq:appendix_heatvarying}
\end{equation}
where $a(x,y)$ is prescribed, $u(x,y,0)=0$, and
$u|_{\partial[0,1]^2}=0$.

\subsection{KS 1D}
\label{app:ks}
For $(x,t)\in[0,2\pi]\times[0,1]$,
\begin{equation}
u_t+\alpha uu_x+\beta u_{xx}+\gamma u_{xxxx}=0,\qquad
(\alpha,\beta,\gamma)=
\left(\frac{100}{16},\frac{100}{16^2},\frac{100}{16^4}\right),
\label{eq:appendix_ks}
\end{equation}
with $u(x,0)=\cos(x)[1+\sin(x)]$.

\subsection{NS 2D CG}
\label{app:nsbackstep}
Let $\Omega=([0,4]\times[0,2])\setminus([0,2]\times[1,2])$.
For velocity $\boldsymbol{v}=(u,v)$ and pressure $p$,
\begin{equation}
(\boldsymbol{v}\cdot\nabla)\boldsymbol{v}
+\nabla p-\nu\Delta\boldsymbol{v}=\boldsymbol{0},\qquad
\nabla\cdot\boldsymbol{v}=0,\qquad \nu=0.01.
\label{eq:appendix_nsbackstep}
\end{equation}
At the inlet $(u,v)=(4y(1-y),0)$; on solid walls $(u,v)=(0,0)$;
at the outlet $p=0$.

\subsection{Poisson 2D C}
\label{app:poissonclassic}
Let $\Omega=[-0.5,0.5]^2\setminus\bigcup_{j=1}^{4}D_j$, where
$D_j$ have radius $0.1$ and centers $(\pm0.3,\pm0.3)$. Then
\begin{equation}
\Delta u=0\quad\text{in }\Omega,\qquad
u=1\quad\text{on the outer boundary},\qquad
u=0\quad\text{on hole boundaries}.
\label{eq:appendix_poissonclassic}
\end{equation}

\subsection{Poisson 2D CG}
\label{app:poissonboltzmann}
The domain $[-1,1]^2$ has four circular holes with $(x,y,r)$ values

$$
(0.5,0.5,0.2),\quad(0.4,-0.4,0.4),\quad
(-0.2,-0.7,0.1),\quad(-0.6,0.5,0.3).
$$

The PDE is
\begin{equation}
\begin{aligned}
-\Delta u+64u&=f(x,y),\\
f(x,y)&=10(17+x^2+y^2)\sin(\pi x)\sin(4\pi y),
\end{aligned}
\label{eq:appendix_poissonboltzmann}
\end{equation}
with $u=0.2$ on the outer boundary and $u=1$ on hole boundaries.

\subsection{Poisson 2D MA}
\label{app:poissonmanyarea}
On $\Omega=[-10,10]^2$,
\begin{equation}
a(x,y)\Delta u+f(x,y)=0,\qquad \partial_nu+u=0
\quad\text{on }\partial\Omega.
\label{eq:appendix_poissonmanyarea}
\end{equation}
For a $5\times5$ grid of cells, $a(x,y)=a_{ij}$ and
\begin{equation}
f(x,y)=F_{ij,00}
+\sum_{m,n=0}^{1}F_{ij,mn}
\sin\!\left(\frac{m\pi\xi}{4}\right)
\sin\!\left(\frac{n\pi\eta}{4}\right),
\label{eq:appendix_poissonmanyarea_force}
\end{equation}
where $(\xi,\eta)$ are local cell coordinates and $a_{ij},F_{ij,mn}$
are prescribed coefficients.

\subsection{Poisson 3D CG}
\label{app:poisson3d}
The domain is $[0,1]^3$ with four spherical holes specified by

$$
(0.4,0.3,0.6,0.2),\quad(0.6,0.7,0.6,0.2),\quad
(0.2,0.8,0.7,0.1),\quad(0.6,0.2,0.3,0.1),
$$

where each tuple gives $(x,y,z,r)$. The equation is
\begin{equation}
-\mu(z)\Delta u+k(z)^2u=f(x,y,z),\qquad
\mu(z)=1,\quad
k(z)=\begin{cases}8,&z<0.5,\\10,&z\geq0.5,\end{cases}
\label{eq:appendix_poisson3d}
\end{equation}
with
\begin{equation}
\begin{aligned}
f(x,y,z)={}&20\exp\!\left[
\sin(\pi x)+\sin(10\pi y)+\sin(5\pi z)\right]
\frac{x^2+y^2+z^2-1}{x^2+y^2+z^2+1}\\
&+100\sin(\pi x)\sin(10\pi y)\sin(5\pi z).
\end{aligned}
\label{eq:appendix_poisson3d_force}
\end{equation}
All boundaries satisfy $\partial_nu=0$.

\subsection{Poisson 5D}
\label{app:poisson5d}
For $\boldsymbol{x}\in[0,1]^5$,
\begin{equation}
\Delta u+\frac{\pi^2}{4}\sum_{j=1}^{5}
\sin\!\left(\frac{\pi x_j}{2}\right)=0,\qquad
u|_{\partial\Omega}=\sum_{j=1}^{5}
\sin\!\left(\frac{\pi x_j}{2}\right).
\label{eq:appendix_poisson5d}
\end{equation}
The analytic solution is
$u^*(\boldsymbol{x})=\sum_{j=1}^{5}\sin(\pi x_j/2)$.

\section{Implementation and Training Details}
\label{app:implementation}

All compared methods use a budget of 20,000 optimizer steps per run.
PINNMorph maintains a population of $K=5$ adaptation policies. The same initial architecture,
initialization, training data, and optimizer-step budget are used for
PINNMorph and its ablation variants.

\subsection{Base PINN and Optimization}

The base model is a four-hidden-layer MLP of width 128 with \texttt{tanh}
activation. The \texttt{burgers\_2d} model additionally uses periodic
positional encoding. Training uses Adam with learning rate $10^{-3}$ followed
by an L-BFGS phase with learning rate $1.0$, without a learning-rate
scheduler.

\subsection{Online Adaptation Schedule}

The training state is diagnosed every 200 optimizer steps. Structural
intervention is enabled from step 400 and disabled after step 14,000, leaving
the final stage for optimization of the resulting model. Following a
structural intervention, the warm-up consists of 10 activation-only steps
followed by 50 steps that optimize newly introduced parameters. Effects are
evaluated at $+100$, $+500$, and $+1000$ optimizer steps, corresponding to
short-, medium-, and persistent-horizon feedback. A new intervention is not
initiated while the previous intervention is awaiting persistent feedback.

\subsection{Decision and Audit Sets}
\label{app:probes}

Two held-out collocation sets, $\mathcal{B}_{\mathrm{dec}}$ and
$\mathcal{B}_{\mathrm{aud}}$, are sampled independently before training and
remain fixed. Neither contributes gradients to PINN optimization. The
decision set contains up to 128 points per physical component and at most
4,096 points in total, subject to the corresponding training-sampling budget.
The audit set contains up to 2,048 points per component and at most 32,768
points in total. The decision set supports online diagnosis and action
selection; the separate audit set checks intervention outcomes and
checkpoint quality. Reference solutions are not used by either set.

\subsection{State Transfer}

Structural interventions modify the PINN already under optimization.
Unchanged parameters and compatible optimizer states are retained. New
parameters are initialized by the registered morphing operator and added to
the active optimizer. Appendix~\ref{app:action_transfer} gives the transfer
and preservation checks.

\subsection{LLM Configuration}

Policy generation and evolution use DeepSeek-V4-Flash. Candidate scoring
uses Qwen3.8-27B. PINNsAgent also uses DeepSeek-V4-Flash.
For PINNMorph, the configured temperature is $0.7$ for policy evolution and
$0.0$ for candidate scoring. The generation model produces structured action
requests, the scoring model returns candidate scores, and deterministic code
validates and executes each selected intervention. The same LLM configuration
is used across the PDE problems.

\section{Online Action Space and Deterministic Execution}
\label{app:action_space}

\subsection{Action space}
\label{app:action_space_catalogue}

The policy language has two levels. A policy first chooses a high-level action
$a\in\mathcal A$ and a compatible mechanism $m\in\mathcal M(a)$; an ordered
operation list, if supplied, gives the authoritative execution request. The
current executable registry contains 16 high-level actions (including
\textsc{Keep}) and 23 primitive mechanisms. Table~\ref{tab:action_space}
groups the primitives into six \emph{descriptive} families. These six groups
are a presentation device, rather than six unrestricted choices in the JSON
schema: action--mechanism--target compatibility is checked before execution.
The same configured primitive library is available for every PDE; the current
diagnostic state determines which actions are applicable.

\begingroup
\footnotesize
\setlength{\tabcolsep}{3pt}
\renewcommand{\arraystretch}{1.18}
\begin{longtable}{@{}>{\RaggedRight\arraybackslash}p{0.115\linewidth}>{\RaggedRight\arraybackslash}p{0.205\linewidth}>{\RaggedRight\arraybackslash}p{0.405\linewidth}>{\RaggedRight\hyphenpenalty=10000\exhyphenpenalty=10000\arraybackslash}p{0.205\linewidth}@{}}
\caption{Executable online mechanisms. ``Zero output'' means that the added
module leaves the current function unchanged at insertion. Numeric arguments
are optional; the executor derives defaults from the validated strength and
diagnostic state.}\label{tab:action_space}\\
\toprule
Group & Mechanism & Target and execution rule & Parameters \\
\midrule
\endfirsthead
\multicolumn{4}{l}{\emph{Table~\ref{tab:action_space} continued}}\\
\toprule
Group & Mechanism & Target and execution rule & Parameters \\
\midrule
\endhead
\bottomrule
\endfoot
Topology & Widen & Replicate units in selected hidden layers and divide outgoing weights by replication count (Net2Wider map). & Width fraction; layers. \\
Topology & Regrow & Expand selected hidden layers through the same function-preserving duplication map. & Width fraction; layers. \\
Topology & Add layer & Insert a new hidden layer after a selected layer; initialize its weights and continue the live model. & Layer index. \\
Topology & Prune & Remove low-utility hidden units while respecting a minimum width. & Width fraction; minimum width; layers. \\
Topology & Redistribute & Remove donor units and add capacity elsewhere in the hidden stack. & Move fraction; minimum width; layers. \\
Topology & Local reinitialization & Reinitialize a bounded fraction of units in selected hidden layers. & Reset fraction; layers. \\
Topology & Component capacity reallocation & Prune shared width and add a zero-output expert for a chosen output component, subject to a parameter-count tolerance. & Component; branch size; tolerance. \\
Additive & Residual branch & Add a gated correction of the current output, initialized at zero. & Branch width and depth; target. \\
Additive & Correction branch & Add a zero-output input-to-output corrector in parallel with the incumbent. & Branch width and depth. \\
Additive & Fourier branch & Feed a fixed, axis-balanced sine/cosine frequency bank to a zero-output corrector. & Branch size; frequency scale. \\
Additive & Gated branch & Mix two encoded input streams with learned feature gates; the final output map starts at zero. & Branch width and depth. \\
Additive & Localized experts & Add overlapping, residual-centered experts with zero output maps and spatial softmax windows. & Expert count; width; overlap; centers. \\
Additive & Adaptive Fourier branch & Add a zero-output Fourier corrector with trainable frequencies and optional localization. & Branch size; scale; bandwidth; region. \\
Additive & Component expert & Add a zero-output corrector for one output component only. & Component; branch size. \\
Additive & Split conflicting topology & Add a gated component-specific zero-output branch selected from conflict diagnostics. & Component; branch size. \\
Objective & Rebalance loss weights & Rescale registered PDE/BC/IC weights toward a common component-gradient norm. & Weight bounds; strength. \\
Objective & Balance component gradients & Apply bounded weight balancing and enable periodic component-gradient normalization. & Weight bounds; update rate. \\
Objective & Reweight residual regions & Upweight a selected component and introduce pointwise weighting in high-residual, boundary, or interface regions. & Weight bounds; update rate. \\
Objective & Emphasize constraints & Multiply BC/IC terms upward and PDE terms downward within bounded ratios. & BC/IC and PDE scales. \\
Objective & Mitigate gradient conflict & Apply the same bounded component-weight response when the diagnosed issue is gradient conflict. & BC/IC and PDE scales. \\
Gradient & Project conflicting gradients & Set the strength of the registered pairwise gradient-projection rule. & Projection strength. \\
Sampling & Adapt sampling distribution & Replace a fraction of mutable collocation points using residual scores and exploratory candidates; keep the point budget fixed. & Replace share; explore share; pool size. \\
Optimizer & Enter L-BFGS phase & Request entry into the configured later L-BFGS phase when an eligible Adam plateau is detected. & Plateau length; latest step. \\
\end{longtable}
\endgroup

The structural macro-actions are \textsc{Conservative Growth},
\textsc{Exploratory Morph}, \textsc{Reallocation},
\textsc{Representation Augmentation}, \textsc{Local Specialization},
\textsc{Spectral Adaptation}, \textsc{Component Capacity Adaptation}, and
\textsc{Gradient Topology Adaptation}. The five objective macro-actions are
\textsc{Objective Rebalancing}, \textsc{Constraint Emphasis},
\textsc{Gradient Conflict Mitigation}, \textsc{Region Aware Rebalancing}, and
\textsc{Gradient Surgery}. \textsc{Sampling Adaptation} and
\textsc{Optimization Phase Switch} complete the intervention menu;
\textsc{Keep} is an explicit no-op. Each macro-action has a registered subset
of the mechanisms in Table~\ref{tab:action_space}. For example,
\textsc{Conservative Growth} admits widening, regrowth, or a residual branch;
\textsc{Local Specialization} admits localized experts; and
\textsc{Region Aware Rebalancing} admits component-gradient balancing or
residual-region reweighting. A multi-operation program must use a macro-action
compatible with every listed primitive; operations execute in list order as
one transaction.

\paragraph{Parameter domains.}
The schema restricts every non-\textsc{Keep} strength to $(0,1]$ and its
target selector to \textsc{All}, \textsc{HighestUtility},
\textsc{LowestUtility}, \textsc{Widest}, or \textsc{Narrowest}. Width, reset,
move, replacement, and exploration fractions are in $(0,1]$; direct hidden
layers retain at least two units. Additive branch widths are at least four
and depths at least one. Fourier scales lie in $[10^{-4},10^4]$; localized
expert overlap is in $[0.1,10]$, and its residual and distance powers are
in $[0.1,8]$. Explicit gradient-projection strengths and objective
adaptation rates lie in $(0,1]$. The executor enforces further
capability-specific constraints before changing the live model.

\subsection{Structured policy output and execution semantics}
\label{app:action_execution}

The LLM emits a declarative policy, not PyTorch code. Its action request is
$a_t=(a,m,\gamma,\sigma,O)$, where $a$ is a macro-action, $m$ a registered
mechanism or \textsc{Auto}, $\gamma=(\text{region},\text{selector})$ a target,
$\sigma\in(0,1]$ a strength, and $O$ an optional ordered primitive list.
The executor validates the fields and translates the request into numeric
operations using the live model and reference-free diagnostics:
\begin{equation}
  \text{LLM policy}\ \longrightarrow\
  \text{validated action}\ \longrightarrow\
  \text{deterministic execution plan}\ \longrightarrow\
  \text{continued optimization}.
  \label{eq:action_execution_pipeline}
\end{equation}
The LLM neither supplies executable training code nor directly writes model
weights or optimizer tensors. This representative \emph{valid} policy payload
requests a single function-preserving intervention:
\begin{quote}
\footnotesize
\begin{verbatim}
{
  "policy_spec": {
    "conditions": {"optimization.plateau": true},
    "program": {
      "diagnosis": "capacity bottleneck",
      "trigger": {"plateau_steps": 600},
      "action": "CONSERVATIVE_GROWTH",
      "mechanism": "WIDEN",
      "target": {"region": "middle_layers",
                 "selector": "highest_utility"},
      "strength": 0.2,
      "operations": [{"primitive": "WIDEN", "strength": 0.2,
        "parameters": {"width_fraction": 0.2}}],
      "recovery_steps": 500,
      "cooldown_steps": 800,
      "fallback": {},
      "reason": "Plateau with stable gradients."
    }
  },
  "capability_gap_proposals": []
}
\end{verbatim}
\end{quote}
Here \texttt{operations} is authoritative. Newly generated non-\textsc{Keep}
policies are instructed to provide at least one operation. An empty list is
accepted for historical action-only checkpoints, in which case the executor
chooses a compatible primitive by a fixed state-dependent rule; it is not an
invitation for the LLM to invent an implementation.

For a residual branch, the actual additive map is
\begin{equation}
  \widetilde u(x)=u_{\theta}(x)
     +\alpha\,g_{\phi}\!\left(u_{\theta}(x)\right),
  \qquad \alpha\big|_{\rm insertion}=0.
  \label{eq:action_residual_branch}
\end{equation}
Correction, Fourier, gated, and component branches instead add a
zero-output map $g_{\phi}(x)$ or $g_{\phi}(\Phi(x))$, where $\Phi$ is the
corresponding feature transform. A localized expert bank adds
\begin{equation}
  \widetilde u(x)=u_{\theta}(x)
   +\sum_{j=1}^{K}\omega_j(x)g_{\phi_j}(x),\qquad
  \omega_j(x)=\frac{\exp[-\tfrac12\|(x-c_j)/s_j\|_2^2]}
  {\sum_{\ell=1}^{K}\exp[-\tfrac12\|(x-c_\ell)/s_\ell\|_2^2]},
  \label{eq:action_local_experts}
\end{equation}
where centers $c_j$ are selected from residual-evaluation points and every
expert's final map starts at zero. The objective action for component $k$
uses its measured gradient norm $G_k$ to update
\begin{equation}
  w_k^+=\operatorname{clip}_{[w_{\min},w_{\max}]}
  \left[w_k\,\operatorname{clip}_{[r_{\min},r_{\max}]}
  \left(\frac{\operatorname{median}_j G_j}{G_k}\right)^{\eta}\right],
  \label{eq:action_loss_reweight}
\end{equation}
with bounded $\eta$ and ratios. Sampling actions replace
$\lceil\rho N\rceil$ points from a mutable set of size $N$ with scored
high-residual and exploratory candidates; the set size remains $N$.
For a conflicting pair of component gradients $g_i,g_j$ with
$g_i^{\top}g_j<0$, gradient surgery applies the bounded projection
\begin{equation}
  g_i\leftarrow g_i-
  \lambda\frac{g_i^{\top}g_j}{\|g_j\|_2^2+\varepsilon}g_j,
  \qquad \lambda\in(0,1],
  \label{eq:action_gradient_projection}
\end{equation}
before summing component gradients. The registered L-BFGS action requests
the next configured optimizer phase; it does not create another model.

\subsection{Parameter and optimizer-state transfer}
\label{app:action_transfer}

Let $A_t$ denote the live architecture and $\theta_t$ its parameters at a
decision step. A committed structural action applies a deterministic map
\begin{equation}
  (A_t,\theta_t,S_t)\xrightarrow{\;T(a_t)\;}
  (A_{t}^{+},\theta_{t}^{+},S_t^{+})
  \xrightarrow{\;\text{next optimizer step}\;}
  (A_{t}^{+},\theta_{t+1},S_{t+1}),
  \label{eq:action_transfer}
\end{equation}
where $S_t$ is the optimizer state. No new training trajectory is launched.
Unchanged parameters retain their values and parameter identities, and
compatible optimizer moments are retained. When a parameter tensor is
resized, the executor maps its surviving entries and corresponding optimizer
state tensors to the replacement parameter. Newly introduced parameters are
initialized by the registered operator and added to the optimizer; their
moments are initialized lazily. Other modules, loss terms, sampled points,
and counters remain in place unless they are the target of the action.

The zero-output additive branches and Net2Wider-style growth preserve the
model function at insertion, up to floating-point error. The trainer checks
the output, selected first and second derivatives, and registered PDE
residuals on the fixed held-out decision and audit sets before accepting such a morph. Reduction
or reinitialization operators (pruning, redistribution, component-capacity
reallocation, adding a direct hidden layer, and local reinitialization) are
\emph{not} asserted to be exact
function-preserving maps; they modify the incumbent in place, keep compatible
state, and undergo finite-output/derivative/residual checks followed by the
same recovery and feedback process. If execution or a required preservation
check fails, the pre-action model and runtime snapshot are restored. The
default post-morph optimizer-state mode is preservation; an explicit reset
configuration is a distinct option. A later Adam-to-L-BFGS transition follows
the configured optimizer phases and does not restart the network.

\subsection{Validation, rejection, and no-op decisions}
\label{app:action_validation}

The policy parser limits conditions to registered state paths, checks the
macro-action and mechanism enumerations, enforces compatible target regions,
and rejects unknown parameters or out-of-range numeric values. In particular,
non-\textsc{Keep} strengths lie in $(0,1]$; recovery and cooldown are bounded
at 2{,}000 and 5{,}000 steps. The executor then checks the currently enabled
primitive set, model compatibility, parameter budget, availability of the
decision and audit sets, and remaining optimization budget. An unavailable
primitive may be replaced only by a registered, target-compatible equivalent,
with the substitution logged; otherwise the candidate is rejected. An ordered
operation list is atomic: failure of any operation rolls back the sequence.
\textsc{Keep} is selected when no candidate is applicable or a lifecycle gate
blocks intervention; it has no executable operations and leaves the training
trajectory unchanged.

\section{Counterfactual Continuation and Execution Feedback}
\label{app:counterfactual_feedback}

\paragraph{Frozen continuation estimate.}
At intervention step $t_0$, each set independently fits a forecast from its
last $W=7$ pre-intervention observations $(t_i,\ell_{k,i})$ for every held-out
physical loss component $k$. The same fit is made for the total score and for spatial
regions when their set of region labels is stable. With $\epsilon=10^{-12}$,
$t_{\rm ref}=t_W$ (the latest recorded step), and
$y_{k,i}=\log(\ell_{k,i}+\epsilon)$, the
robust log-linear fit is
\begin{align}
s_k &= \operatorname{median}_{i<j}
       \frac{y_{k,j}-y_{k,i}}{t_j-t_i}, \notag\\
b_k &= \operatorname{median}_i
       \bigl[y_{k,i}-s_k(t_i-t_{\rm ref})\bigr], \notag\\
\widetilde\ell^0_k(t_0+h)
    &= \max\!\left\{0,
       \exp\!\left[b_k+s_k(t_0+h-t_{\rm ref})\right]-\epsilon\right\}.
\end{align}
Each forecast is fixed at $t_0$ and is not refitted using post-intervention
measurements. This is a local extrapolation of recent set behavior; no
unmodified shadow training trajectory is run.

\paragraph{Noise control and eligibility.}
For each fitted metric, let $m_k=\operatorname{median}_i
|y_{k,i}-b_k-s_k(t_i-t_{\rm ref})|$ and
$D=\max(\max_i t_i-\min_i t_i,1)$. We clip the raw forecast
$\widetilde\ell^0_k$ between $\ell_{k,\min}/4$ and
$\max(4\ell_{k,\max},\epsilon)$ to obtain $\widehat\ell^0_k$.
Its confidence at horizon $h$ is
\begin{equation}
c_{k,h}=\exp\!\left[-1.4826m_k
 \left(1+\frac{|t_0+h-t_{\rm ref}|}{D}\right)\right],
\label{eq:continuation_confidence}
\end{equation}
multiplied by $0.25$ if the envelope bound changes the prediction.
Confidence is combined by geometric mean first within each active physics
group and then across groups, including the region group when available.
Direct credit at a horizon requires both sets to have confidence at least
$0.35$ and no clipped prediction. Fewer than seven observations, duplicate
steps, changed component sets, or non-finite fitted values invalidate the
forecast. A mismatch between predicted and observed region sets withholds
credit. Forecast ineligibility does not by itself prevent a safe intervention;
it withholds numerical policy credit. Although the trigger configuration specifies a
minimum of three observations, the physics-progress gate uses the larger of
that value and the seven-point continuation window. It therefore uses seven
observations under the reported configuration.

\paragraph{Per-horizon utility and cross-set rule.}
For each set $\mathcal B$ and component $k$, the continuation-relative
utility at horizon $h$ is
\begin{equation}
u^{\mathcal B}_{k,h}=
\frac{\widehat\ell^{0,\mathcal B}_k(t_0+h)-
      \ell^{\mathcal B}_k(t_0+h)}
     {\max\{\widehat\ell^{0,\mathcal B}_k(t_0+h),\epsilon\}}.
\label{eq:continuation_component_utility}
\end{equation}
Let $\mathcal G$ contain the active PDE, boundary, initial, and observation
groups. For each group $g$, let $\bar u^{\mathcal B}_{g,h}$ be the mean of its
component utilities. When spatial regions are available, the least improved
region enters $\mathcal G$ as one additional group. All active groups have
equal weight, giving
\begin{equation}
U_h^{\mathcal B}=\frac{1}{|\mathcal G|}
  \sum_{g\in\mathcal G}\bar u^{\mathcal B}_{g,h}.
\label{eq:continuation_group_utility}
\end{equation}
A set accepts an outcome if its
aggregate utility is at least $0.01$ and no individual component or region
degrades by more than $0.10$ relative to its forecast. Both sets must
accept for positive credit, and the credited utility is the smaller of
their two aggregate utilities. If either set is ineligible, or the two sets
show opposite global or focused worst-region directions, numerical credit
is withheld. Opposite directions mean one utility is above $10^{-12}$ and
the other is below $-10^{-12}$, checked both globally and for either set's
worst-region identifier. Agreed degradation is eligible for negative credit. All
outcomes remain in the shared execution history; same-direction but
low-confidence outcomes can contribute only bounded qualitative evidence
($|\mathrm{hint}|\leq 0.25$) to later policy evolution.

\paragraph{Three distinct feedback events.}
The $+100$, $+500$, and $+1000$ observations are recorded separately under
the intervention identifier and horizon. There is no weighted sum across
horizons. Each unconfounded, credit-eligible event updates the executed
policy's credit using
$C_n=C_{n-1}+(r_n-C_{n-1})/n$, where $n$ counts that policy's credited
horizon outcomes. Here $r_n$ is the magnitude of the smaller set utility,
with a positive sign for accepted outcomes and a negative sign for failures.
A failed safety guard is classified as failure even if aggregate utility is
positive. These forecasts reduce contamination by the
trajectory's recent trend, but remain estimated counterfactuals rather than
observed causal controls.

\bibliographystyle{plainnat}
\bibliography{references}
\end{document}